\documentclass[conference]{IEEEtran}
\usepackage{cite}
\usepackage{amsmath,amssymb,amsfonts}
\usepackage{algorithmic}
\usepackage{graphicx}
\usepackage{textcomp}
\usepackage{xcolor}
\def\BibTeX{{\rm B\kern-.05em{\sc i\kern-.025em b}\kern-.08em
    T\kern-.1667em\lower.7ex\hbox{E}\kern-.125emX}}

\usepackage{url}
\usepackage{hyperref}
\usepackage{subcaption}
\usepackage{authblk}

\begin{document}

\title{Determining Standard Occupational Classification Codes from Job Descriptions in Immigration Petitions}

\author[1]{Sourav Mukherjee}
\author[2]{David Widmark}
\author[2]{Vince DiMascio}
\author[3]{Tim Oates}
\affil[1]{Fairleigh Dickinson University, Vancouver, Canada. Email: sourav.mukherjee.8@gmail.com}
\affil[2]{Berry Appleman \& Leiden LLP, USA. Email: \{dwidmark, Vdimascio\}@balglobal.com}
\affil[3]{Synaptiq, USA. Email: tim.oates@synaptiq.ai}

\maketitle

\begin{abstract}

Accurate specification of standard occupational classification (SOC) code is critical to the success of many U.S.~work visa applications. Determination of correct SOC code relies on  careful study of job requirements and comparison to definitions given by the U.S.~Bureau of Labor Statistics, which is often a tedious activity. In this paper, we apply methods from natural language processing (NLP) to computationally determine SOC code based on job description. We implement and empirically evaluate a broad variety of predictive models with respect to quality of prediction and training time, and identify models best suited for this task.
\end{abstract}

\section{Introduction}
\label{sec:Introduction}

The process of obtaining U.S.~work visas has become increasingly difficult in recent years. Data from the U.S.~Citizenship and Immigration Services (USCIS) \cite{uscis-rfe-stats} show that between Fiscal Year (FY) 2015 and FY 2019, the approval rate of H-1B visa petitions has declined substantially. In response to some (though not all) visa petitions, the USCIS requests further information by issuing a Request For Evidence (RFE) \cite{RFE}. Per \cite{uscis-rfe-stats}, the percentage of petitions that resulted in RFEs has increased dramatically between FY 2015 and FY 2019, reaching 40.2\% in 2019. Further, the percentage of petitions with RFE that were approved has decreased significantly during this time frame, reaching a low of 62.4\% in FY 2018. In FY 2020, approval rates showed some improvement but were still lower than those in FY 2015; similarly, RFE issuance rate somewhat decreased but was still higher than that in FY 2015 \cite{bal-rfe-stats}. We note that issuance of RFE, petitioner's response, and subsequent review of response by the USCIS, add significant delays even for petitions that are ultimately approved. Per the USCIS, the most common reason for RFE issuance is the petitioner (employer) not being able to establish that the position is a \textit{specialty occupation} \cite{RFE}  (please refer to \cite{H1B} and \cite{SpecialtyOccupation} for explanations of this term). Therefore, accurate characterization of job positions is critical to the successful and timely completion of the visa petitions.

The U.S.~Bureau of Labor Statistics (BLS) has created the Standard Occupational Classification (SOC) \cite{SOC-1} to categorize jobs into 867 occupational categories. Each category is denoted by an SOC code. The mapping of SOC codes to categories is given in \cite{SOC-2}. To minimize the chances of a visa petition getting delayed (due to RFE) or denied, it is important that the petitioner specify the SOC code that best describes the duties associated with the position. Typically, this requires careful reading of the job description and SOC code definitions to find the best match. Although SOC codes are organized hierarchically to facilitate search, the process is tedious, and results in enormous repetitive workload for immigration law firms.

In this paper, we focus on the problem of algorithmically determining SOC codes based on job descriptions. Applying techniques from natural language processing (NLP), we build a variety of predictive models that accept free form textual descriptions of job duties as input, and yield SOC code as output. Using real world data, we empirically evaluate these models with respect to quality of prediction and training time, and identify models that are best suited for this task.

The rest of this paper is organized as follows. Section \ref{sec:RelatedWork} reviews related work. In Section \ref{sec:Methodology}, we describe our approach. Section \ref{sec:Evaluation} presents our experimental evaluation which are interpreted in Section \ref{sec:Discussion}. Finally, Section\ref{sec:Conclusion} summarizes the paper and discusses future directions.

\section{Related Work}
\label{sec:RelatedWork}

Machine learning methods have been applied in the past to various problems in the legal domain \cite{Surden2014,10.1007/978-3-030-19823-7_31,Faggella2020}, such as outcome forecasting \cite{10.2307/4099370,10.2307/3688543,Katz2016,Aletras2016PredictingJD,Medvedeva2019UsingML}, document discovery \cite{DBLP:conf/icail/YangGFY17,Cormack2014EvaluationOM}, document categorization \cite{Lemley2007,DBLP:conf/bigdataconf/WeiQYZ18,Silva2018DocumentTC,DBLP:conf/fedcsis/UndaviaMO18,DBLP:conf/cikm/LuCAK11,DBLP:conf/propor/FurquimL12,Kumar2012}
and legal drafting \cite{DBLP:conf/afips/SprowlBCEK84,Betts2017,Miller}.

Application of machine learning methods to immigration law is a much newer area of research. The problem of predicting outcomes of refugee claims has been considered in \cite{DBLP:conf/icail/DunnSSC17,DBLP:conf/icail/ChenE17}. In contrast, our paper focuses on work visa applications as opposed to refugee claims, and seeks to programmatically select SOC codes as opposed to predicting case outcomes.

In \cite{DBLP:conf/icdm/MukherjeeODJAWH20}, two problems related to work visa applications are considered, namely, categorization of supporting documents of visa petitions, and drafting responses in reaction to Requests For Evidence (RFE). Our work is different from \cite{DBLP:conf/icdm/MukherjeeODJAWH20} in that we focus on identifying SOC codes programmatically in an effort to proactively reduce the chances of RFE issuance.

Interestingly, application of natural language processing to determine SOC code has been studied in the epidemiological context. Specifically, the SOCcer (Standardized Occupation Coding for Computer-assisted Epidemiologic Research) model \cite{pmid27102331} predicts SOC code based on industry, job title, and job tasks. Our work is different from SOCcer in the following ways. First, while SOCcer is trained and evaluated using health-related datasets, we focus on data related to work visa petitions. Second, while SOCcer uses an ensemble of classifiers, three of which are based on job title, one on industry, and one on task, we seek to predict SOC code using description (i.e., tasks and responsibilities) alone. This is due to our observation that in work visa related data, job titles do not map to SOC codes in a consistent way and that the number of distinct SOC codes associated with an industry such as the software industry is huge. Third, unlike SOCcer, we benchmark a broad variety of models and compare them in terms of accuracy and training time. Finally, our benchmarking includes two different text vectorization approaches, namely sparse vectorization (using TF-IDF $n$-grams  \cite{DBLP:conf/ecml/Joachims98}) and dense vectorization (using doc2vec \cite{DBLP:conf/icml/LeM14}, a neural network).     

The next section describes our approach in detail.

\section{Methodology}
\label{sec:Methodology}

We begin by formally defining our problem.  
\subsection{Problem Statement}
\label{subsec:ProblemStatement}

\noindent \textbf{Input:}
\begin{enumerate}
    \item $\Sigma$: a finite alphabet. $\Sigma^+$ denotes the set of all non-empty strings over $\Sigma$. In this paper, we focus on strings that are job descriptions expressed as free form text.
    \item $\mathcal{Y}$: a finite set of labels. In this paper, SOC codes are treated as labels.
    \item $\mathcal{D} = \{(x_i, y_i): 1 \leq i \leq n \}$: a labeled dataset of size $n \in \mathbb{N}$, where $x_i \in \Sigma^+$ is a job description, and $y_i \in \mathcal{Y}$ is its corresponding SOC code.
\end{enumerate}

\noindent \textbf{Output:}
A function $f: \Sigma^+ \rightarrow \mathcal{Y}$ which maps a job description $x$ to an SOC code $y = f(x)$ such that $f$ minimizes the expected error with respect to some loss function.

From a pragmatic standpoint, we want such a function $f$ to be available as a web service (i.e., web API) which accepts a request containing description $x$ to produce a response containing the predicted SOC code $y = f(x)$.

\subsection{Approach}
\label{subsec:Approach}

Our approach may be described as a sequence of steps as follows.

\subsubsection{Text Vectorization}
Since a majority of machine learning algorithms assume inputs to be real valued vectors, predictive modeling based on text often requires vectorizing the text, i.e., computing real valued vector representation of text. We consider two different vectorization techniques, which are as follows.
\paragraph{TF-IDF $n$-grams} An $n$-gram ($n \in \mathbb{N}$) is a sequence of $n$ tokens. Given $n_{\mathrm{min}}, n_{\mathrm{max}} \in \mathbb{N}$ ($n_{\mathrm{min}} \leq n_{\mathrm{max}}$), a corpus of text in $\Sigma^+$ can be used to compute the vocabulary of all $n$-grams where $n_{\mathrm{min}} \leq n \leq  n_{\mathrm{max}}$. Subsequently, any string $x \in \Sigma^+$ may be represented as a vector of counts, i.e., term frequencies (TF) of $n$-grams present in $x$. Such a vector representation of a string is typically sparse, i.e., most of its components are zero, since most $n$-grams in the vocabulary are typically absent in it. To offset the effect of highly frequent $n$-grams with little semantic value, the vectors are weighted by inverse document frequencies (IDF), resulting in TF-IDF $n$-gram representations.
While TF-IDF representations have been found to achieve high accuracy in text categorization \cite{DBLP:conf/ecml/Joachims98}, the high dimensionality of the sparse vectors generally entails high computational costs for training predictive models.
\paragraph{Doc2vec} An alternative approach that addresses the issue of dimensionality consists of using neural architectures for vectorizing words \cite{mikolov2013efficient} and strings \cite{DBLP:conf/icml/LeM14}, using contextual similarity to predict semantic similarity. The resulting representations are known as word embeddings and document embeddings, respectively, and the above neural architectures are referred to as word2vec and doc2vec, respectively. Embeddings computed by word2vec and doc2vec are typically of lower dimensionality compared to TF-IDF $n$-gram representations. Therefore, such embeddings are considered dense vector representations. Since job descriptions are strings of arbitrary length, we use doc2vec to compute dense vector representations of such descriptions.

\subsubsection{Predictive Modeling}
For each type of vectorization, we train a set of standard classifiers for predicting SOC code, namely, $k$-nearest neighbors (KNN), Gaussian na\"ive Bayes (GNB), logistic regression (LR), linear support vector machine (LinearSVC), support vector machine with radial basis function (SVC-RBF), decision tree (DT), and random forest (RF).

\subsubsection{Evaluation and Model Selection}
To evaluate the models, we use $n$-fold cross validation. The dataset is first divided into $n$ slices (or folds) of (roughly) equal size. In each round of cross validation, a different slice is held out for testing while the remaining $n - 1$ slices are used for training. Several metrics are recorded in each round. At the end of $n$ rounds of training and testing, these metrics are averaged and reported. These scores help identify models best suited to the problem.

\subsubsection{Deployment}
Once a model has been selected, we deploy it as a web service which can accept a \texttt{POST} request whose body contains a job description in free form text and produce a response containing the predicted SOC code.

The next section presents our empirical evaluation.

\section{Evaluation}
\label{sec:Evaluation}

\subsection{Dataset}

Our dataset consists of 46,999 labeled instances, where each instance corresponds to a visa petition. For every instance, the relevant attributes include \textit{job title}, \textit{job description}, \textit{company name}, \textit{SOC code (normalized)} (which we will refer to as simply \textit{SOC code}), and \textit{SOC occupation} which is a moniker of the SOC code. We exclude company name from the model since we have found it to be irrelevant to the predictive task; moreover, the predictive model should be able to generalize to all companies. We exclude job title from the model as well because we have found many instances of the same job title being associated with different SOC codes in this dataset, suggesting that job title does not consistently map to SOC code. Therefore, we use job description as the only input to our models. Since SOC occupation is simply a moniker of SOC code, we use SOC code as the only output of our models.

It is worth noting that the distribution of SOC codes in this dataset is uneven. While the dataset includes abundant examples of the most common categories, less frequent codes may not have sufficient instances. To build a predictive model that is accurate for a majority of use cases, we focus on the 5 most frequent codes, which results in a dataset with 32,262 instances.

\subsection{Experimental Setup}

Our experiments are implemented using Python 3 as the programming language, in interactive notebooks hosted on the Databricks\footnote{\url{https://databricks.com/}} platform. Other standard libraries used include Scikit-learn \cite{scikit-learn} for sparse vector representations and training classifiers, Gensim for doc2vec  \cite{rehurek_lrec}, Numpy \cite{numpy} for numerical computations, Pandas \cite{pandas} for tabular data processing, and Matplotlib \cite{matplotlib} for plotting. We use Managed MLflow\footnote{https://databricks.com/product/managed-mlflow} for deployment.

We have implemented and benchmarked 14 classifiers, 7 of which are based on TF-IDF $n$-gram representation, while the rest are based on doc2vec representation. These are compared in terms of training time, accuracy, precision, recall, and f1 score \cite{scikit-learn-metrics}.  The values of these metrics are averaged over 10-fold cross validation and reported.

\subsection{Hyperparameters}

All hyperparameters used in this evaluation are manually tuned. Automatic parameter tuning is outside the scope of this paper and left as future work.

\subsubsection{Vector Representation}

For TF-IDF $n$-gram representation, we use $ 1 \leq n \leq 10$. However, $n$-grams that occur in fewer than 10\% of the instances or greater than 90\% of the instances are ignored. The resulting sparse vectors have a dimensionality of 858. For doc2vec, we use a dimensionality of 100.

\subsubsection{Predictive Modeling}

For $k$-nearest neighbor classifiers, we use $k = 3$. For random forest classifiers, we use an ensemble of 100 estimators.

\subsection{Experimental Results}

\subsubsection{Accuracy}

\begin{figure}[h!]
    \centering
    \includegraphics[scale = 0.50]{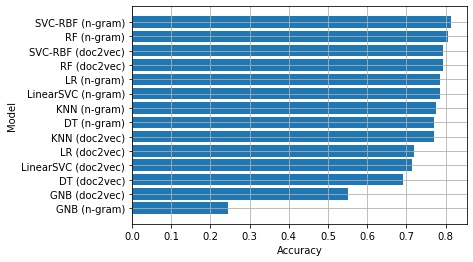}
    \caption{Accuracy scores of SOC Code predictors.}
    \label{fig:Accuracy}
\end{figure}

We measure accuracy as the fraction of predictions that are correct. Figure \ref{fig:Accuracy} compares the accuracies of the models being evaluated.

\subsubsection{Precision}

\begin{figure}[h!]
    \centering
    \includegraphics[scale = 0.5]{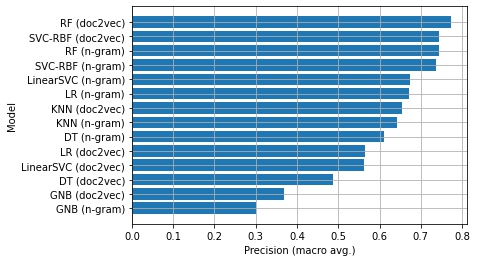}
    \caption{Precision (macro average) of SOC Code predictors.}
    \label{fig:Precision}
\end{figure}

In a binary classification problem, precision is defined as the fraction of all positive predictions that are correct. Since our problem involves more than two classes, we report the macro average, i.e., the average of precision scores measured with respect to each SOC code in the dataset \cite{scikit-learn-precision}. Figure \ref{fig:Precision} shows the macro average precision scores of the models.

\subsubsection{Recall}

\begin{figure}[h!]
    \centering
    \includegraphics[scale = 0.5]{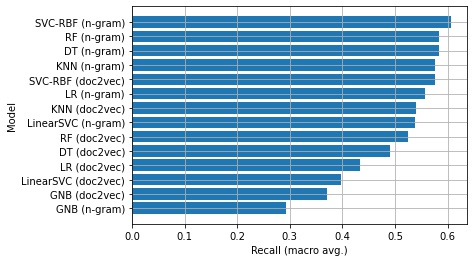}
    \caption{Recall (macro average) of SOC Code predictors.}
    \label{fig:Recall}
\end{figure}

In a binary classification problem, recall is defined as the fraction of all positive instances that are correctly predicted as positive. Since our problem involves more than two classes, we report the macro average, i.e., the average of recall scores measured with respect to each SOC code in the dataset \cite{scikit-learn-recall}. Figure \ref{fig:Recall} shows the macro average recall scores of the models.

\subsection{F1 Score}

\begin{figure}[h!]
    \centering
    \includegraphics[scale = 0.5]{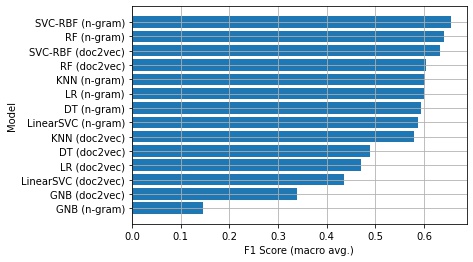}
    \caption{F1 score (macro average) of SOC Code predictors.}
    \label{fig:f1}
\end{figure}

In a binary classification problem, f1 score is defined as the harmonic mean of precision and recall. Since our problem involves more than two classes, we report the macro average, i.e., the average of f1 scores measured with respect to each SOC code in the dataset \cite{scikit-learn-f1}. Figure \ref{fig:f1} shows the macro average f1 scores of the models.

\subsection{Training Time}

\begin{figure}[h!]
    \centering
    \includegraphics[scale = 0.5]{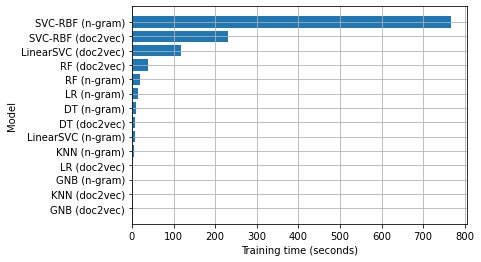}
    \caption{Training time of SOC Code predictors.}
    \label{fig:TrainingTime}
\end{figure}

Finally, the time taken (in seconds) to train each model (averaged over 10-fold cross validation as will all the other metrics) is shown in Figure \ref{fig:TrainingTime}.

In the next section, we interpret these results.

\section{Discussion}
\label{sec:Discussion}

Figure \ref{fig:Accuracy} shows that TF-IDF $n$-gram based support vector classifier with radial basis function (SVC-RBF) achieves the highest classification accuracy of 0.813031. We also note that regardless of whether the text vectorization is sparse (TF-IDF $n$-grams) or dense (doc2vec), SVC-RBF and random forest classifiers achieve high accuracy in the neighborhood of approximately 0.8. 

Figures \ref{fig:Accuracy}, \ref{fig:Recall}, and \ref{fig:f1} further indicate that TF-IDF $n$-gram based SVC-RBF achieves the highest cross validation scores with respect to accuracy, recall, and f1 score. However, Figure \ref{fig:Precision} shows doc2vec based random forest achieves the highest precision score.

These results demonstrate that while support vector classifiers with radial basis functions and random forest classifiers are suitable models for SOC code prediction, the choice of representation (sparse vs.~dense) may depend on the metric of highest importance.

Figure \ref{fig:TrainingTime} shows that the high accuracy of TF-IDF $n$-gram based SVC-RBF comes at the cost of high training time, which is significantly greater than all the other models considered in this study. On the other hand, doc2vec based SVC-RBF requires much lower training time and yet achieves comparable accuracy. We note that the dimensionality of the sparse vectors is 858 while that of the dense vectors is 100, which is likely a contributing factor to this disparity in training time. We further observe that random forest classifiers, whether based on sparse or dense vectors, can be trained even more quickly while still achieving comparable accuracy.

Therefore, in a real world deployment, the choice of model may be dependent on the trade-off between training time and accuracy. Let us consider a scenario where once an initial model has been deployed, more accurate models are trained in the background as more training data become available over time, allowing the web service to switch to such models when they are substantially more accurate. If there are time constraints associated with the initial deployment, random forest or doc2vec based SVC-RBF would provide a highly accurate model more quickly. Subsequently, if there are no time constraints on switching to newer models, then TF-IDF $n$-gram based SVC-RBF may be preferable for later deployments. The next section concludes the paper.

\section{Conclusion}
\label{sec:Conclusion}

Accurate determination of Standard Occupational Classification (SOC) codes is critical to the success and timely completion of U.S.~work visa applications. In this paper, we have applied machine to reduce the repetitive workload of SOC code selection. Using methods from natural language processing, we have trained a variety of predictive models for determining SOC code based on job description. Using real world data, we have benchmarked these models with respect to quality of prediction and training time. Our results indicate that our approach results in highly accurate models that may be trained and deployed within reasonable timelines.

Several useful extensions of this work are possible. For example, the functionality of the models may be expanded to return a list of suggested SOC codes ranked by some confidence metric. Another improvement would be to incorporate statistical significance tests (e.g., Student's t-test) into the model comparison process.

\bibliographystyle{IEEEtran}
\bibliography{mybib}

\end{document}